\documentclass[conference]{IEEEtran}
\IEEEoverridecommandlockouts
\usepackage{cite}
\usepackage{amsmath,amssymb,amsfonts}
\usepackage{algorithmic}
\usepackage{graphicx}
\usepackage{textcomp}
\def\BibTeX{{\rm B\kern-.05em{\sc i\kern-.025em b}\kern-.08em
    T\kern-.1667em\lower.7ex\hbox{E}\kern-.125emX}}
\usepackage{amsmath}
\usepackage{amssymb}
\usepackage{multirow}
\usepackage[xcdraw]{xcolor}
\usepackage[colorlinks=true, allcolors=blue]{hyperref}
\usepackage{glossaries}
\definecolor{darkblue}{rgb}{0.0, 0.0, 0.5}

\usepackage{pgfplots}
\usepackage{tikz}
\usetikzlibrary{shapes.geometric, arrows}

\newacronym{cdf}{CDF}{cumulative density function}
\newacronym{sa}{SA}{survival analysis}
\newacronym{kmeier}{KM}{Kaplan-Meier}
\newacronym{coxph}{CoxPH}{\mbox{Cox proportional hazard model}}
\newacronym{rsf}{RSF}{random survival forest}
\newacronym{atf}{ATF}{accelerated time-failure model}
\newacronym{weibullatf}{WATF}{Weibull accelerated time-failure model}
\newacronym{boostedcox}{CBoost}{\mbox{gradient boosting with CoxPH loss function}}
\newacronym{ci}{CI}{concordance index}
\newacronym{ir}{IR}{isotonic regression}
\newacronym{ibs}{IBS}{integrated Brier score}
\newacronym{mape}{MAPE}{mean absolute percentage error}
\newacronym{cpp}{CPP}{Canon Production Printing}
\newacronym{ml}{ML}{machine learning}

\begin{document}

\title{Predicting the Lifespan of Industrial Printheads with Survival Analysis}

\author{
\IEEEauthorblockN{Dan Parii, Guangzhi Tang,\\ Charalampos S. Kouzinopoulos, Marcin Pietrasik}
\IEEEauthorblockA{\textit{Department of Advanced Computing Sciences} \\
\textit{Maastricht University}\\
Maastricht, The Netherlands }
\and
\IEEEauthorblockN{Evelyne Janssen}
\IEEEauthorblockA{\textit{Quality Processes and Validation Department} \\
\textit{Canon Production Printing}\\
Venlo, The Netherlands }}

© 2025 IEEE. Personal use of this material is permitted. Permission from IEEE must be
obtained for all other uses, in any current or future media, including
reprinting/republishing this material for advertising or promotional purposes, creating new
collective works, for resale or redistribution to servers or lists, or reuse of any copyrighted
component of this work in other works.

This paper has been published in the \textit{8th IEEE Conference on Industrial Cyber-Physical Systems (ICPS)} in Emden, Germany, May 12-15, 2025.

D. Parii, E. Janssen, G. Tang, C. Kouzinopoulos and M. Pietrasik, ``Predicting the Lifespan of Industrial Printheads with Survival Analysis,'' \textit{2025 IEEE 8th International Conference on Industrial Cyber-Physical Systems (ICPS)}, Emden, Germany, 2025, pp. 1-6, doi: 10.1109/ICPS65515.2025.11087918.


\maketitle

\begin{abstract}
Accurately predicting the lifespan of critical device components is essential for maintenance planning and production optimization, making it a topic of significant interest in both academia and industry. In this work, we investigate the use of survival analysis for predicting the lifespan of production printheads developed by Canon Production Printing. Specifically, we focus on the application of five techniques to estimate survival probabilities and failure rates: the Kaplan-Meier estimator, Cox proportional hazard model, Weibull accelerated failure time model, random survival forest, and gradient boosting. The resulting estimates are further refined using isotonic regression and subsequently aggregated to determine the expected number of failures. The predictions are then validated against real-world ground truth data across multiple time windows to assess model reliability. Our quantitative evaluation using three performance metrics demonstrates that survival analysis outperforms industry-standard baseline methods for printhead lifespan prediction.
		
\end{abstract}


\section{Introduction}
\label{intro}

Printheads are vital printer components responsible for transferring ink or toner onto paper during the printing process. As such, their performance directly affects print quality and costs related to refurbishment, recalls, and customer retention. As a printer manufacturer, \gls*{cpp} is focused on maintaining high standards, regular evaluation, and oversight of printheads that are already deployed in the field. One aspect of this is the desire to gain insights into the lifetime distribution of their printheads. Previous efforts to estimate the lifespan of printheads used the \gls*{kmeier} model, generating a single failure rate for the entire population. This approach, however, proved inaccurate with prediction errors significantly exceeding a desired $10\%$ limit. This provides the motivation for our work wherein we investigate the use of \gls*{sa} as an improved prediction model to predict the lifespan of printheads.

Widely used in the field of predictive maintenance \cite{soares-2002}, \gls*{sa} describes a set of methods for analyzing time-to-event data, focusing on estimating the time until an event, such as printhead failure, occurs \cite{clark2003survival}. It is important to note that it accounts for \textit{censored observations}, data points where the event of interest has not yet been observed within the study period. By accounting for these type of observations, \gls*{sa} provides more accurate and unbiased estimates of survival probabilities and hazard rates, preventing distortion in the analysis. 
This is particularly relevant to our work, where more than $70\%$ of printhead data is censored \cite{turkson2021handling}.

Due to the limited number of studies on predicting failure rates, we conducted a comprehensive evaluation of \gls*{sa} methodologies. Specifically, our work explores a diverse selection of models spanning the four classes described by Wang et al. \cite{wang2019machine}: non-parametric, semi-parametric, fully-parametric, and \gls*{ml}-based techniques. While these approaches share the common goal of modeling time-to-event data, they rely on different underlying assumptions. Specifically, non-parametric models such as \gls*{kmeier}, require only historical failure data, making them flexible but potentially less informative when covariates are available. In contrast, semi-parametric models like the \gls*{coxph} impose strict data assumptions such as proportional hazards and linear relationships between covariates and the log-hazard function. While these assumptions enhance interpretability, violations can lead to biased estimates.

Fully-parametric models, such as the accelerated failure time model, require accurate assumptions on the distribution of the survival times. In our work, we assume a Weibull distribution, commonly used in survival data \cite{carroll2003use}. \gls*{ml} models such as random survival forests \cite{ishwaran2008random} and gradient boosting \cite{friedman2001greedy} are chosen to uncover complex, non-linear relationships in the data.
Once survival probabilities were obtained by these methods, calibration using \gls*{ir} \cite{ma2014identifying} was introduced to address a possible under-prediction bias due to heavy data censoring \cite{jiang2022novel}.

To approximate the number of failing printheads from our estimated survival probabilities, we considered each printhead as a Bernoulli variable. As such, to obtain the expected number of values, we calculate the expected value $E(X)$ for all random variables, where the failure probability is determined by our SA models. This approach was evaluated on real-world data obtained from CPP and a quantitative evaluation demonstrated the viability of all investigated techniques, with \gls*{kmeier} achieving the best performance. 

\section{Related Works}\label{litreview}



The field of \gls*{sa} can be broadly divided into two categories of approaches: statistical inference and \gls*{ml}. In our review of the literature, we focus on these categories separately before briefly highlighting common calibration methods. 

Statistical methods are rooted in probability theory and traditional inference techniques. Canonical examples include the \gls*{kmeier} estimator, the \gls*{coxph} model \cite{cox1972regression}, and accelerated failure time models \cite{wei1992accelerated}. In Wang et al. \cite{wang2019machine}, the Weibull distribution predicts battery cell lifespan with limited features, emphasizing feature selection. 
Closer to our work, Snider and McBean \cite{snider2021combining} compare random forests, \gls*{rsf} with a Weibull proportional hazard model in predicting the lifespan of water mains, with \gls*{rsf} emerging as the superior model.
Moat and Coleman \cite{moat2021survival} use \gls*{kmeier}, \gls*{coxph}, and \gls*{atf} models to predict the remaining lifetime of boilers, concluding that maintenance factors do not play as big of a role as the production date of the boiler and when it was installed.

In survival data, censored observations pose challenges for standard \gls*{ml} methods as they are akin to unlabeled samples in classification or unknown responses in regression \cite{wang2019machine}. Unlike unlabeled samples, however, censored instances provide partial information indicating the possible range of the true response (survival time). This partial information must be carefully handled within any \gls*{ml} method to ensure accurate predictions and has been explored in the literature. For instance, the random survival forest modifies the splitting criteria from class purity, as used in the original random forest method \cite{breiman2001random} to the ordering of survival times, thereby including censored instances. This model then creates simple, featureless, and counting-based estimators such as KM or the Nelson-Aalen \cite{nelson1972theory} estimators for more homogeneous populations in the leaf nodes. The approach addresses the limitation of models like \gls*{kmeier}, which provide a single distribution for the entire population by generating more accurate estimates for subpopulations.
Papathanasiou et al. \cite{papathanasiou2023machine} leverage an \gls*{rsf} model on a synthetic dataset for the purpose of predictive maintenance.
Gradient boosting can be extended to handle censored data by incorporating a loss function such as the aforementioned CoxPH loss function to yield \gls*{boostedcox} \cite{binder2008allowing} and using the estimated covariates to derive the survival function.
SurvivalSVM \cite{van2008survival} is an extension of the widely used support vector machine (SVM) model, which alters the optimization objective to maximize the correct ordering of individuals. This method is primarily useful for ranking individuals rather than generating survival probabilities.
Neural networks have also been leveraged for SA as in Biganzoli et al. \cite{biganzoli1998feed} wherein a partial logistic artificial neural network demonstrated the use of neural approaches for survival data, especially in high-dimensional settings. 


Calibration methods such as \gls*{ir} \cite{ma2014identifying}, can be used to adjust predicted probabilities that fall outside the [0, 1] range. Niculescu-Mizil and Caruana \cite{niculescu2005obtaining} compared various scaling methods applied to a boosting-based model, their findings showed that IR significantly improved cross-entropy and yielded superior mean squared error results. Similarly, Berta et al. \cite{berta2024classifier} showed that combining logistic regression with IR effectively aligns model outputs with actual probabilities.

\section{Methodology}\label{methods}

\begin{figure}
    \centering
    \includegraphics[width=0.95\linewidth,height = 0.5\linewidth]{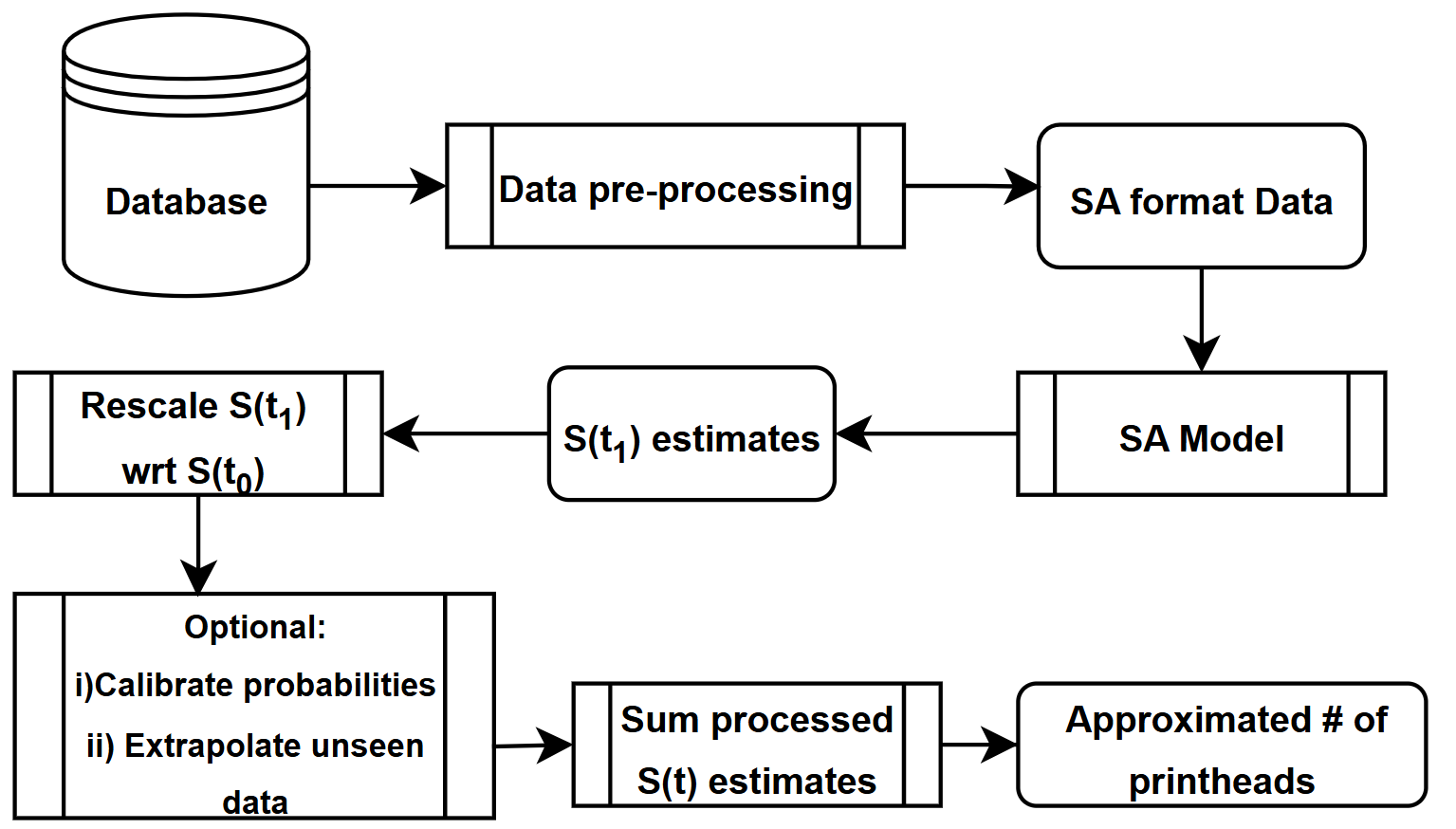}
    \caption{Outline of the prediction workflow as described in Section \ref{methods}.}
    \label{fig:predictionworkflow}
\end{figure}

In this section, we first provide an overview of the data and its inherent challenges before introducing the foundations of \gls*{sa}. This is followed by a discussion of the models explored and metrics used to evaluate them.
A graphical summary of our integrated workflow is provided in Figure \ref{fig:predictionworkflow}.

\subsection{Data}
\label{data_section}

The dataset utilized for predictive modeling consists of historical data of printheads, manufactured between 2008 and 2024. While the dataset includes a substantial number of printheads, the available feature set is relatively limited. Notably, the specific printhead model under study was introduced in 2009, before the implementation of nozzle logging, which describes logging that tracks printhead activations and movement on a granular level. This type of data, now widely used in more recent models, would have provided valuable insights but is unavailable for the earlier printheads.

The available data is derived from two primary logging sources, printer information and the printhead logging mechanism. The printer information provides metadata on printhead position, managed color, active regions, and installation dates. Conversely, the printhead logging mechanism captures operational data, including \textit{Warm Hours} (total active time) and the volume of jetted toner ink.

\subsubsection{Data Challenges}

As mentioned in Section \ref{intro}, the dataset is \textit{heavily} censored, with $70\%$ of the observations incomplete. This is primarily due to many printheads still being operational, making their time-to-event unknown. Additionally, the logging data is often unreliable due to irregular logging frequencies between printheads. Furthermore, a limited feature set leads to similar printheads having different lifespans, suggesting unmeasured factors that affect their performance. As such, predicting failure within a specific time \(t\) is challenging, and attempts to classify failure per printhead have been inaccurate. The irregular frequency of logging limits the possibility of robust time-series analysis. Finally, domain experts at \gls*{cpp} noted the potential of erroneous data. Specifically, certain printheads may show inflated usage data due to their installation in printers not connected to the main servers, resulting in discrepancies between usage data and other recorded features.

\subsubsection{Data Cleaning}\label{subsub:datacleaning}

In order to address some of these challenges, data cleaning was performed. Specifically, printheads with excessively large usage statistics were removed with thresholds obtained through consultation with \gls*{cpp} domain experts. Furthermore, old printheads and those with highly uncommon time-in-use (more than 12 hours a day) were excluded. Additionally, printheads stored for more than 1.5 years between production and first installation were marked with a boolean indicator as they were likely installed in unconnected printers beforehand and then moved to connected ones. Furthermore, dead-on-arrival printheads (those that had some defect from the start and failed quickly) were removed, as they are a small and unrepresentative subset of the data. 
We note that despite efforts to eliminate unreliable printheads, some may still be present in the cleaned data due to their usage statistics falling within the expert-defined thresholds.

\subsection{Survival Analysis}
\label{sub:survivalanalysis}

In this section, we introduce the fundamental notations and terminologies pertinent to \gls{sa}, along with an overview of the approaches employed in our work. 
As mentioned earlier, the goal of \gls{sa} is to estimate and analyze the time until an event of interest occurs while accounting for censored data and identifying factors that influence survival probabilities.
To formalize this, we will use the notation of Wang et al. \cite{wang2019machine}.
Thus, we represent a given instance \(i\) by the triplet \((X_{i}, y_{i}, \delta_{i})\), where \(X_{i} \in \mathbb{R}^{1 \times P}\) is the feature vector; \(\delta_{i}\) is the Kronecker delta binary event indicator with  $\delta_{i} = 1$ for an uncensored instance and $\delta_{i} = 0$ for a censored instance); and \(y_{i}\) denotes the observed time. It equals the survival times \(T_{i}\) and \(C_{i}\) for uncensored and censored instances, respectively.
	\[
	y_i =
	\begin{cases} 
		T_i & \text{if } \delta_i = 1, \\
		C_i & \text{if } \delta_i = 0.
	\end{cases}
	\]
The goal of \gls{sa} is to estimate \(T_{i}\) for a new instance, with features from \(X_{j}\), presenting a traditional challenge with the caveat of taking \(C_{i}\) records into account.

The main estimate of \gls*{sa} is the survival function which represents the probability that the time to the event of interest is not earlier than a specified time, $t$. The survival function is represented by a \gls*{cdf} of $S(t)$, given as: 
		\[
		S(t) = \Pr(T \geq t)
		\]
Where \(S(t)\) is a monotonically decreasing function of time \(t\) in the range $[0,1]$ and represents the probability of survival beyond time \(t\).
The survival function CDF \( S(t) \) provides an estimate of the survival probability at a given time \( t \). \( S(t) \) represents the probability that the machine has not failed by time \( t \).
To approximate the survival function as a probability of failure within a specific time interval, we need to scale the estimates based on the assumption that the event of interest has not occurred at the starting time \( i \). For the survival function \( S(t_{i,j}) \), we assume that \( S(i) = 1 \), as the machine has not failed by time \( i \). Given this, using a derivation of Bayes Rule, we can express $S(t_{i,j})$ as:
 \[
 \text{Assuming } i \leq j, \quad
 S(t_{i,j}) = 
 \begin{cases} 
     1, & \text{if } i = j, \\
     \frac{S(j)}{S(i)}, & \text{if } i \neq j.
 \end{cases}
 \]
To calculate the expected number of failures by time \(j\), we compute \(F(t_{i,j})\), the inverse of \(S(t_{i,j})\), for each individual, generating a set of failure probabilities. Each individual is modeled as a Bernoulli variable, where failure is the event of interest and the probability \(p\) is the model's output. The expected number of failures is then the sum of these probabilities:
 \[
 E(X) = \sum_{i=1}^{n} p_i
 \]

\subsection{Survival Analysis Models}

To generate the survival estimates we will employ five different models: \gls*{kmeier}\cite{kaplan1958nonparametric}, \gls*{coxph}, \gls*{rsf}, \gls*{boostedcox} and the \gls*{weibullatf}. Below is a brief overview of these models.

\subsubsection{\gls*{kmeier}}

The \gls*{kmeier} model is a lightweight non-parametric method that discretizes survival data into bins, then counts the failure rate for each group and approximates a distribution with a respective survival probability for each probability bin. So, the approximation is only based on a counting process of failures. 
Specifically, let \(T_{1}<T_{2}<T_{3}..,T_{K-1}<T_{K}\), be a set of ordered event times for $N$ instances. For a specifc event-time \(T_{j}\), the number of events are \( d_{j} \geq 1 \) and \(r_{j}\) are individuals at risk since their event time is higher than \(T_{j}\). Using this terminology, we can calculate the conditional probability of surviving beyond \(T_{j}\) as $p(T_j) = \frac{r_j - d_j}{r_j}$.
From this we derive the survival function \(S(t) = P(T \ge t)\) :	
		\[S(t) = \prod_{j : T_j < t} p(T_j) = \prod_{j : T_j < t} \left( 1 - \frac{d_j}{r_j} \right)\]
The \gls*{kmeier} model is simple but offers key advantages: it handles missing attributes, is computationally efficient, and is more robust to erroneous data. 

\subsubsection{\gls*{coxph}}

As a semi-parametric model, \gls*{coxph} does not assume a distribution of the survival times. The model does, however, rely on the proportional hazard assumption which assumes that the hazard function, \(h(t)\), representing the instantaneous risk of failure is constant over time for different individuals when adjusted for their covariates. In other words, the relative risk between individuals remains the same throughout the study period. This assumption allows for a comparison of hazard rates across different groups while accounting for other variables. 
The assumption is formulated as:  
     \[h(t, X_i) = h_0(t) \exp(X_i \beta)\]
Where \(h_{0}(t)\) is the baseline function; \(X_{i} = (x_{i1},x_{i2},.., x_{iK})\) is the covariate vector for instance $i$ and \({\beta}^T = (\beta_1, \beta_2, \dots, \beta_P)\) are the coefficients that need to be estimated. The model is semi-parametric since we do not need to assume the baseline function \(h_{0}(t)\). Based on the assumption, we can then calculate the survival function as: 	
		\[S(t) = \exp\left(-H_0(t) \exp(X \beta)\right) = S_0(t) \exp(X \beta)\]
Where \((S_0 = \exp(-H_0(t)))\) denotes the baseline survival function.
This model is selected as it is particularly effective and reliable in visualizing the impact of covariates. 

\subsubsection{Gradient Boosting CoxPH}

An extension of \gls*{coxph}, \gls*{boostedcox} uses gradient boosting \cite{friedman2001greedy} tailored for survival data. It combines weak learners, specifically regression trees, sequentially such that each new tree corrects the errors of the previous ones by minimizing the negative gradient of the partial likelihood from \gls*{coxph}. This iterative process improves the model's predictive performance for survival outcomes by capturing complex, non-linear relationships in time-to-event data while maintaining the proportional hazards assumption.
However, the \gls{boostedcox} model risks overfitting and tends to underperform on noisy data. 

\subsubsection{Random Survival Forest}

The random forest algorithm \cite{breiman2001random} is an ensemble learning method that relies on weak learners. Specifically, it constructs decision trees during training before aggregating their predictions to improve accuracy and reduce overfitting.
The \gls*{rsf} \cite{ishwaran2008random} applies this approach to survival data by using survival trees instead of decision trees. Specifically, instead of traditional classification or regression splits, it utilizes survival-specific splitting criteria, such as maximizing the log-rank statistic, to partition the data based on differences in survival distributions. The final prediction is obtained by aggregating the cumulative survival probabilities from individual trees with a simple estimate such as the \gls*{kmeier} or Nelson-Aalen \cite{nelson1972theory} estimators.

\subsubsection{Accelerated Time-Failure Model}

The \gls*{atf} model is the sole parametric method implemented in this study. The model assumes a specific parametric distribution for the survival times and estimates the corresponding parameters to model time-to-event outcomes \cite{lee2003statistical}. As such, a key aspect of utilizing \gls*{atf} is selecting the appropriate distribution. To this end, tests were conducted to fit the data to candidate distributions, with the Weibull distribution providing the best fit compared to alternatives as assessed by the Kolmogorov-Smirnov test. The formulation results in the \gls*{weibullatf} model, the most commonly used \gls*{atf} model in survival analysis.
The \gls*{weibullatf} model assumes linearity between the logarithm of survival time and the covariates:
		\[\ln(T) = X \beta + \sigma \epsilon\]
Where $X$ denotes the covariates, \(\beta\) represents the coefficient vector and \(\delta\) is a scaler for the error variable \(\epsilon\), which has the same distribution \(\text{ln}(T)\). 
Next, the selected error distribution is estimated using maximum likelihood estimation. The coefficients $\beta$ and $\delta$ are estimated using standard numerical optimization methods.
In comparison to the aforementioned models, the \gls*{weibullatf} model offers high interpretability as well as the potential for higher performance if the chosen distribution corresponds well to the data.

\subsection{Evaluation Metrics}

Three evaluation metrics were leveraged to assess our methodology: the \gls*{ci}, \gls*{ibs}, and \gls*{mape} of the amount predictions. The first two metrics are tailored for evaluating \gls*{sa} models as they account for censoring. The former measures how well individuals are ranked based on their survival times and functions as a discrimination metric, whereas the latter evaluates how well predicted survival probabilities align with actual outcomes and functions as a calibration metric.
The \gls*{ci} calculates the proportion of correctly ordered pairs of records relative to the total number of pairs \cite{harrell1996multivariable}. This ordering-based metric ensures that censored observations contribute information when comparing survival times. The \gls*{ci} is given by: 
\[
CI = \frac{\sum_{i<j} \mathbb{I}( \hat{T}_i > \hat{T}_j \text{ and } T_i > T_j )}{\sum_{i<j} \mathbb{I}(T_i \neq T_j)}
\]
Where \( T_i \) and \( T_j \) are the true event times for individuals \( i \) and \( j \), respectively, and \( \hat{T}_i \) and \( \hat{T}_j \) are the predicted event times for those individuals. The indicator function \( \mathbb{I}(\cdot) \) is 1 if the condition inside is true, and 0 otherwise.
The \gls*{ibs} is essentially the mean squared error of predicted probabilities adjusted for time-to-event data \cite{gerds2006consistent}. The score is derived from the original Brier Score, which measures the accuracy of probabilistic predictions: 
\[
{BS}(t) = \frac{1}{N} \sum_{i=1}^{N} \left( \hat{S}_i(t) - Y_i(t) \right)^2
\]
Where \( N \) is the number of individuals, \( \hat{S}_i(t) \) is the predicted survival probability for individual \( i \) at time \( t \), and \( Y_i(t) \) is the event indicator for individual \( i \) at time \( t \) (1 if the event occurred, 0 if censored). This evaluates how well a model performs when predicting for one time point $t$. To evaluateover time, we take the integral of the BS function over the specified time range, resulting in the \gls*{ibs}:
\[
	IBS = \frac{1}{t_{\text{max}}} \int_{0}^{t_{\text{max}}} BS(t) \, dt
\]
Where \( t_{\text{max}} \) is the maximum time of interest.
The final metric, \gls*{mape},  measures prediction accuracy by normalizing residuals as a percentage of actual values, providing a straightforward way to assess alignment with true failure counts. We formalize this as follows:
\[
	{MAPE} = \frac{1}{n} \sum_{i=1}^{n} \left| \frac{A_i - F_i}{A_i} \right| \times 100
	\]
Where \(A_i\) is the predicted number of failures, \(F_i\) is the actual number, and \(n\) is the number of predictions.

\subsection{Scaling with Calibration}
\label{sub:other_methods}

As discussed earlier, calibration methods such as \gls*{ir} have been shown to improve model probabilities, particularly in cases with systematic bias. In our case, this bias arises from the heavy censoring of our data. As such, we evaluate our models in both calibrated and uncalibrated settings. 
For calibration, we employ \gls*{ir}, a nonparametric statistical method adapted for calibrating probabilities in binary classification \cite{berta2024classifier, zadrozny2002transforming}.
It predicts probabilities to observed outcomes while preserving the monotonicity of the predictions, ensuring that higher predicted risks correspond to higher empirical event rates.
Following Berta et al. \cite{berta2024classifier}, let \( n \in \mathbb{N}^+ \), \( (p_i, y_i)_{1 \leq i \leq n} \in (\mathbb{R}^2)^n \) be pairs of uncalibrated probabilities and the true labels, and let \( (w_i)_{1 \leq i \leq n} \in (\mathbb{R}^+)^n \) represent a set of positive weights. Assuming the indices are ordered such that \( p_1 \leq p_2 \leq \cdots \leq p_n \), the \gls{ir} problem is defined as:
		\[
		\min_{r \in \mathbb{R}^n} \frac{1}{n} \sum_{i=1}^n w_i (y_i - r_i)^2 \ \ \text{s.t.} \ \   r_1 \leq r_2 \leq \cdots \leq r_n 
		\]
The result of this is an optimization problem, that calculates a non-decreasing piecewise function \( r \), with inputs \( (p_i)_{1 \leq i \leq n} \) that minimizes the squared error with respect to the labels \( (y_i)_{1 \leq i \leq n} \), under a certain weighting \( (w_i)_{1 \leq i \leq n} \) of each data point \( (p_i, y_i)_{1 \leq i \leq n} \). The weight vector is added to give importance to one region of the prediction range if preferred.

\section{Evaluation}

	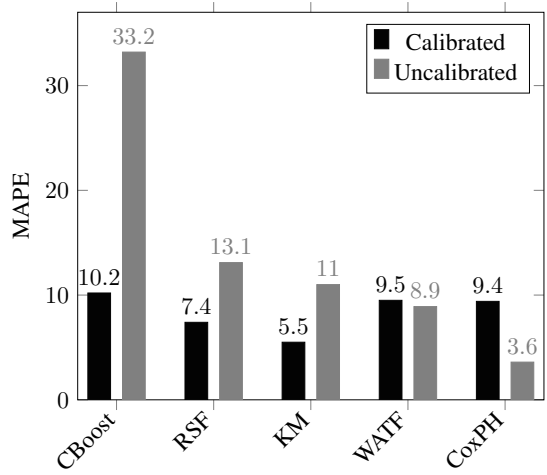
\begin{figure}[t]
	   \centering
		\begin{tikzpicture}[scale=0.9]
        \begin{axis}[
        ybar,
        bar width=9.5pt,
        symbolic x coords={CBoost, RSF, KM, WATF, CoxPH},
        xtick=data,
        ymin=0, ymax=37,
        ybar=5pt,
        ylabel={MAPE},
        xlabel={},
        ylabel near ticks,
        legend pos=north east,
        nodes near coords,
        x tick label style={rotate=45, anchor=east}
    ]
        \addplot[black!99, fill=black!99, single ybar legend,] coordinates {(CBoost,10.2) (RSF,7.4) (KM,5.5) (WATF,9.5) (CoxPH,9.4)};
        \addplot[black!50, fill=black!50,single ybar legend,] coordinates {(CBoost,33.2) (RSF,13.1) (KM,11.0) (WATF,8.9) (CoxPH,3.6)};
        \legend{Calibrated, Uncalibrated}
    \end{axis}
\end{tikzpicture}
		\caption{\gls*{mape} scores of our models in calibrated and uncalibrated settings.}
		\label{fig:bestmapepermodel}
	\end{figure}


To determine the performance of our selected models, we evaluate them across different prediction windows. A prediction window is a specific range of time \(([t_0; t_1])\), where \(t_0\) is a threshold for the maximum date of data log per printhead and \(t_1\) is the future time for which we are predicting.
We use data up to $t_0$ model training and $t_1$ for model evaluation.
We extracted data for six different prediction windows, starting with \([t_0= \text{May 2021} ;t_1=\text{May 2022}]\), and ending with \([t_0= \text{November 2023};t_1=\text{November 2024}]\), such that the difference between the starting points, $t_0$ are six months and their corresponding $t_1$ is one year ahead. 
We evaluated ten configurations —- five models with and without \gls*{ir} calibration —- across six prediction windows with multiple iterations for stable failure probability estimates. The \gls*{mape} is used to determine the best model, calibration, and estimate type. Additionally, we assess the \gls*{ci} and \gls*{ibs} for each model fitted on the latest data to explore their relationship with the \gls*{mape} in predicting failure numbers.


\subsection{Results}


On the whole, the results indicate that \gls*{sa} methods can successfully predict the number of printhead failures. This is highlighted in Figure \ref{fig:bestmapepermodel} which displays the \gls*{mape} of each configuration. We observe that the best configuration for 4 out of 5 models had an average residual percentage error of under $10\%$. The  \gls*{coxph} model had the lowest \gls*{mape} overall with $3.6\%$ for the uncalibrated $F(t)$ estimate, followed by calibrated $F(t)$ for \gls*{kmeier} with a $5.5\%$ error. Calibrated models outperformed uncalibrated ones, except for \gls{coxph} and \gls{weibullatf}. For  \gls{kmeier}, \gls{rsf}, and \gls{boostedcox}, the uncalibrated predictions were less accurate compared to the first two models, and thus, calibration substantially improved their performance. For example, \gls{boostedcox} performed very poorly without calibration, with residual errors reaching up to $33\%$, making it unsurprising that calibration significantly reduced the error.
All models showed similar performance in terms of \gls{ci} scores as shown in Table \ref{tab:SAMetrics}. Ensemble models \gls{boostedcox} and \gls{rsf} performed best with \gls{ci} scores above 0.8, while regression models \gls{coxph} and \gls{weibullatf} obtained slightly lower scores of $0.79$ and $0.77$, respectively. The \gls{kmeier} model was excluded from this comparison as it cannot be evaluated with \gls{ci} due to its uniform survival curve. \gls{ibs} scores were similar across most models, with \gls{boostedcox} performing best at $0.07$. The \gls{kmeier} model performed only slightly better than a random estimator, scoring $0.2$ with $0.25$ expected from random estimation.

\begin{table}[t]
\centering
    \caption{Model results on \gls{ci} and \gls{ibs} using 10-fold evaluation.}
    \begin{tabular}{|l|c|c|}
        \hline
        \textbf{Model}                & \textbf{CI} & \textbf{IBS} \\ \hline
        \gls*{boostedcox}     & \textbf{0.818}             & \textbf{0.077 }                 \\ \hline
        \gls*{rsf} & 0.807             & 0.096                  \\ \hline
        \gls*{weibullatf}           & 0.79              & 0.091                  \\ \hline
        \gls*{coxph}                & 0.774              & 0.094                \\ \hline
        \gls*{kmeier}          & N/A               & 0.2                    \\ \hline
        Random Estimator     & 0.5               & 0.25                   \\ \hline
    \end{tabular}
    \label{tab:SAMetrics}
\end{table}

\subsection{Discussion}

Calibration with \gls*{ir} reduced errors in three out of five models. The \gls*{kmeier} model's \gls*{mape} decreased from $11\%$ to $5.5\%$, indicating calibration’s effectiveness in reducing under-prediction. A more accurate score for this model is more beneficial, as it has the advantage of not relying on missing features.
Two of five models saw slight error increases after calibration, which can be attributed to the variability in risk factors and failure rates across years. For instance, calibrating 2024 data with a model trained on 2023 data may cause overfitting.

The ensemble models \gls*{rsf} and \gls*{boostedcox} performed worse than most of the other methods, with both over-predicting failures. Calibration improved \gls*{boostedcox}'s performance, reducing \gls*{mape} from $33\%$ to $10.2\%$. Despite this improvement, it remains the least effective model. The model’s poor performance may be due to overfitting to the noise in our data
or the way that parameter optimization was performed. Parameter optimization was based on \gls*{ibs}, which doesn’t directly correlate with \gls*{mape} in our six cases, so a good score on the former may prove to overfit predictions for the latter. The relatively better performance of \gls*{rsf} can be attributed to the fact that it is less reliant on clean data, capturing failure rates using the Nelson-Aalen estimator, which is based on failure counting.
We believe simpler models like \gls*{coxph} and \gls*{weibullatf} performed better due to their ability to capture linear relationships, which are less sensitive to noise. The \gls*{kmeier} model also performed well due to the large, consistent dataset, and the fact that the failure rates are stable over time. 

We found no reliable connection between the \gls*{sa} evaluation metrics and our failure approximations. The ensemble models' strong performance on \gls*{ci} and \gls*{ibs}, relative to \gls*{mape} suggests that a high \gls*{ci} does not indicate accurate probability values as it only reflects the correct ordering of predictions. Similarly, \gls*{ibs} evaluates how well a printhead’s survival probability aligns with actual failure rates but it only considers whether a printhead has failed or not, neglecting the full range of probabilities. Therefore, \gls*{ci} and \gls*{ibs} do not directly reflect exact failure probabilities. A more suitable metric for this kind of evaluation might be a calibration curve. This metric gives an overview of how well the predicted probabilities match the actual outcomes. Unlike the binary evaluation approach of the \gls*{ibs}, the calibration curve offers a more nuanced comparison by examining the relationship between predicted failure probabilities and observed failure rates.


\section{Conclusions}

This paper explored the use of \gls*{sa} methods to predict the lifespan of printheads developed by \gls*{cpp}. Specifically, it outlined a methodology for predicting the number of failures using five models from four commonly used classes of methods. These models estimated failure probabilities which were then aggregated to obtain the total number of failures.
The models were evaluated on a real-world dataset obtained from \gls*{cpp} and the results demonstrate the viability of \gls*{sa} methods for lifespan prediction in this context.
Calibration through \gls*{ir} was investigated and proved beneficial in certain models such as \gls*{kmeier}.
In response to the positive results, we aim to extend this approach to other printer models in future work.

\section*{Acknowledgment}

We would like to thank \gls*{cpp} for providing access to data used for model evaluation in this work as well as their guidance and support throughout the research and implementation process.

\bibliographystyle{unsrt}
\bibliography{refs.bib}

\end{document}